\documentclass[letterpaper, 10 pt, journal, twoside]{IEEEtran}
\usepackage{amsmath,amsfonts}
\usepackage{algorithm}
\usepackage{array}
\usepackage[caption=false,font=normalsize,labelfont=sf,textfont=sf]{subfig}
\usepackage{textcomp}
\usepackage{stfloats}
\usepackage{url}
\usepackage{verbatim}
\usepackage{graphicx}
\usepackage{cite}

\usepackage{amssymb}  
\usepackage{enumitem}
\usepackage{algpseudocode}
\usepackage{siunitx}
\usepackage{multirow}
\usepackage{nicematrix}
\usepackage{adjustbox}
\usepackage[font=footnotesize]{caption}
\usepackage{bbm}
\usepackage{soul}
\makeatletter
\let\NAT@parse\undefined
\makeatother
\usepackage{hyperref}
\hypersetup{
  colorlinks=true,
  linkcolor=red,
  urlcolor=blue,
  citecolor=green,
}
\newcommand{\blue}[1]{\textcolor{black}{#1}}
\title{Automated Layout and Control Co-Design of Robust Multi-UAV Transportation Systems}
\author{Carlo Bosio, Mark W. Mueller%
\vspace{-0.5cm}
\thanks{The authors are with the High Performance Robotics Laboratory, University of California, Berkeley. Contact: 
{\tt\small \{c.bosio, mwm\}@berkeley.edu}}%
}
\begin{document}
\maketitle
\begin{abstract}
The joint optimization of physical parameters and controllers in robotic systems is challenging. This is due to the difficulties of predicting the effect that changes in physical parameters have on final performances. At the same time, physical and morphological modifications can improve robot capabilities, perhaps completely unlocking new skills and tasks. We present a novel approach to co-optimize the physical layout and the control of a cooperative aerial transportation system. The goal is to achieve the most precise and robust flight when carrying a payload. We assume the agents are connected to the payload through rigid attachments, essentially transforming the whole system into a larger flying object with ``thrust modules" at the attachment locations of the quadcopters. We investigate the optimal arrangement of the thrust modules around the payload, so that the resulting system achieves the best disturbance rejection capabilities. We propose a novel metric of robustness inspired by $\mathcal{H}_2$ control, and propose an algorithm to optimize the layout of the vehicles around the object and their controller altogether. We experimentally validate the effectiveness of our approach using fleets of three and four quadcopters and payloads of diverse shapes. 
\end{abstract}
\vspace{-0.4cm}
\section{Introduction}
\IEEEPARstart{A}{utonomous} aerial transportation solutions have been proven increasingly essential in applications such as construction, logistics, and load lifting \cite{villa2020survey}. Being able
to scalably and reliably apply aerial robots to these settings would
not only drastically reduce costs and enhance time and energy efficiency, but also reduce the need of ground infrastructures. 
The deployment at scale of Uncrewed Aerial Vehicles (UAVs) to these scenarios is challenging, mainly due to their load capacity and robustness limitations, especially significant when carrying a payload. 
We address the problem of synthesizing a maximally robust multi-UAV system for collaborative payload lifting and transportation. 

A lot of work has been carried out on design, control and path planning for single quadcopter transportation systems, some examples of which are \cite{zeng2020differential, schiano2022reconfigurable, lee2022trajectory, moshref2021applications}. However, practical limitations on dimensions, vehicle
complexity, load capacity, and costs limit the application of such technologies. 
The widespread use and cost-effectiveness of smaller vehicles, such as quadcopters, have made them the preferred choice for practical applications \cite{mark}. 
Their use in a collaborative fashion has been proposed, for instance, for construction applications, but in simplified settings in which each agent individually carries a small load \cite{augugliaro2014flight, real2021experimental}.

Using multiple small vehicles in a cooperative manner allows to increase the payload, but also introduces greater complexity in terms of control, due to e.g.\ aerodynamic interactions and vibrations, and trajectory planning \cite{lee2016planning, lee2018integrated}. Connecting the vehicles to the payload through tethers is a popular choice \cite{estevez2024review}, and allows to distance the agents from the payload. The dynamics and control of tethered systems have been extensively investigated, e.g.\ in \cite{meissen2017passivity, zhang2023formation}. Other examples are \cite{rastgoftar2018cooperative}, which employed an interconnected structure to enhance the system stability, \cite{mohammadi2018decentralized}, which proposed a decentralized approach to the system's control, and \cite{geng2019implementation}, where improved disturbance rejection capabilities were achieved. 
\IEEEpubidadjcol

Compared to the challenges of managing tethered or moving payloads, which can be complex due to their internal dynamics and limited maneuverability, rigid attachments are often the preferred option for transportation purposes. 
Researchers have investigated hardware aspects, such as adding grippers to the vehicles \cite{mellinger2013cooperative}, and control aspects, such as limited sensing capabilities \cite{loianno2017cooperative}, adaptive control frameworks \cite{webb2021adaptive}, and compensation of internal payload vibrations \cite{ritz2013carrying}. 
%
\begin{figure}[tb]
    \centering    \includegraphics[width=0.49\textwidth]{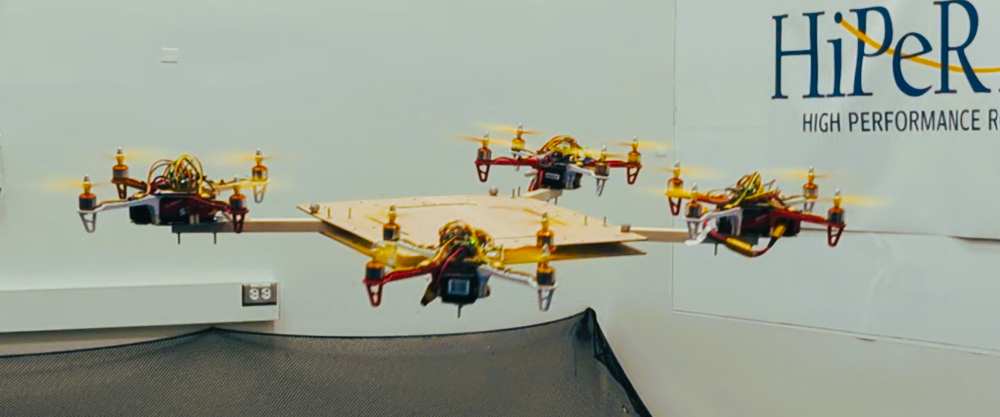}
    \caption{Four quadcopters cooperatively carrying a single panel payload.}
    \label{fig:system_photo}
    \vspace{-0.4cm}
\end{figure}
%
The use of multiple thrust modules \cite{mu2019universal, oishi2021autonomous, chaikalis2023modular} introduces additional parameters to the system, which can be leveraged to enhance performances. The physical configuration of the lifting agents around a payload is particularly interesting, because it can drastically influence the robustness during closed-loop flight. Selecting the best layout, however, can be complex, due to the intricate dependencies of flight performance metrics on physical parameters (which affect the system dynamics, often in a nonlinear fashion). In previous works, such as \cite{mellinger2013cooperative, mu2019universal, webb2021adaptive}, this is done through simple heuristics, or by intuition, \blue{and compensating inaccuracies through adaptive control approaches}, leading to suboptimal flying~systems.

An area addressing such joint control-hardware optimization problems to achieve superior performances is control co-design \cite{garcia2019control}. These methodologies have been successfully applied to a number of robotics applications, such as hands \cite{chen2021co}, legged robots \cite{bravo2022large}, multicopters \cite{carlone2019robot, du2016computational}, and winged drones \cite{zhao2022automatic, bergonti2024co}. The output of these methods is typically a control policy paired with a set of physical system parameters which jointly maximize the desired objective.

Our research applies a co-design approach to the synthesis of collaborative aerial transportation systems. The core idea is to jointly compute the layout and the control of the combined quadcopter-payload system as a single unified problem, maximizing robustness. 
\blue{Our goal is to automate the design process and synthesize systems with high performances by tuning its physical parameters.}
We do this by formulating and solving an optimization problem inspired by robust control theory for a first-order system model, method that has previously been used to capture disturbance rejection performances of a quadcopter \cite{bucki2019novel}. 
\blue{The main contribution of this paper is an efficient co-design tool to optimally arrange thrust modules around a payload to maximize robustness during flight, and consists of two major parts:
\begin{itemize}[leftmargin=*, noitemsep, topsep=0pt]
    \item A novel cost function capturing the disturbance response and the system's input constraints using state-of-the-art methodologies;
    \item A computationally efficient procedure to determine the optimal thrust module arrangement based on the proposed cost function.
\end{itemize}}
%
%
%
%
\vspace{-0.2cm}
\section{Methodology \label{sec:materials}}
%
The co-design problem we solve is intrinsically twofold, as it involves optimizing over a physical configuration as well as over control parameters. The question we ask is: what is the physical placement of $N$ quadcopters around a payload, and associated controller, such that, the overall flying system is the most robust possible to disturbances?
We cast the problem as an optimization problem where the cost function is inspired by $\mathcal{H}_2$ control theory, and the decision variables are the quadcopters' attachment locations. In the inner loop of the optimization, a Linear Quadratic Regulator (or LQR, i.e.\ the $\mathcal{H}_2$-optimal controller) is associated to candidate layouts for evaluation. This choice, in fact, allows a straightforward computation of the control (i.e.\ a linear feedback matrix), and also an analytic evaluation of the $\mathcal{H}_2$-inspired cost function. More advanced control strategies could be used, at the expense of a more involved policy optimization, and a more computationally expensive evaluation of the cost function (using for example Monte Carlo methods). 
We make the assumption that a reliable system for state estimation is available.
\vspace{-0.2cm}
\subsection{Modelling}
\begin{figure}[tb]
    \centering
    \includegraphics[width=0.43\textwidth]{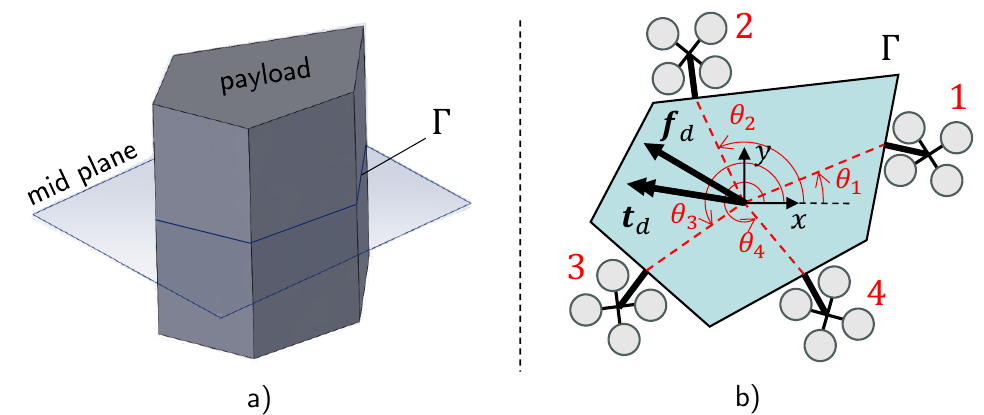}
    \caption{a) Extrusion payload geometry and representation of the mid-plane. Its intersection with the side faces is the curve $\Gamma$. b) Schematics of four quadcopters attached around the payload along $\Gamma$. Representation of the local body reference frame centered at the centroid of $\Gamma$, placement variables defined by the angles $\boldsymbol{\theta_i}$ ($i=1, ..., 4$), and disturbance force $\mathbf{f}_d$ and torque $\mathbf{t}_d$.}
    \label{fig:dronescheme}
    \vspace{-0.4cm}
\end{figure}

We assume that the payload is characterized by a right prism geometry, whose mass properties and shape are known.
These assumptions simplify the following mathematical derivation, but do not limit the space of application. Indeed, many common payloads, such as packages, boxes, pallets all satisfy these assumptions. We investigate the quadcopters' placement on the payload mid-plane (Fig. \ref{fig:dronescheme}a), reducing the amount of decision variables in the optimization problem formulated. 
%
With reference to Fig. \ref{fig:dronescheme}b, the placement variables, are defined as the angles $\boldsymbol{\theta} = [\theta_i, ..., \theta_N]^T$ describing the attachments locations along the curve $\Gamma$ with respect to the centroid of the payload shape. 
The attachment is assumed to be obtained through a rigid rod, which allows to keep distance between payload and thrust~modules. 

The quadcopter-payload ensemble is itself a rigid body, and can therefore be described by a state vector $\mathbf{x} = [\mathbf{p}^T, \Dot{\mathbf{p}}^T, \boldsymbol{\gamma}^T, \boldsymbol{\omega}^T]^T\in \mathbb{R}^{12}$, where $\mathbf{p}$ and $\Dot{\mathbf{p}}$ are the position and velocity of the centroid of the object with respect to a global world reference frame, $\boldsymbol{\gamma}$ is the Roll-Pitch-Yaw triplet defining the orientation of the local frame (this is chosen here for ease of exposition), and $\boldsymbol{\omega}$ the angular velocity with respect to the local body frame. 

Each quadcopter is seen as a thrust module providing four inputs to the system. Therefore, if $N$ quadcopters are deployed, the thrust input vector is $4N$-dimensional, i.e.\ $\mathbf{u}\in\mathbb{R}^{4N}$. We assume that each thrust component is limited to the interval $[u_l, u_h]\subset\mathbb{R}$. 

\blue{For the derivation of the dynamics equations in vector form, we follow \cite{wang2016dynamics}. The motion is described by}

\blue{
\begin{equation}
\begin{cases}
  m \Ddot{\mathbf{p}} = m \mathbf{g} + R \, \mathbf{e}_3 T\\
  J \dot{\boldsymbol{\omega}} + \boldsymbol{\omega}\times J \boldsymbol{\omega}= \boldsymbol{\tau}
\end{cases},
\label{eq:nonlinear-dynamics}
\end{equation}}
\blue{where $m$ and $J$ are the total mass and inertia matrix of the system, $\mathbf{g} = [0, 0, -9.81]^T$ is the gravity acceleration vector, $R$ is the rotation matrix from (3) in \cite{wang2016dynamics} and obtained from the triplet $\boldsymbol{\gamma}$, $\mathbf{e}_3$ is the third vector of the standard Euclidean basis, $T = \sum_{i=1}^{4N} u_i$ is the total thrust produced by all quadcopters, $\mathbf{1}_{4N}$ is the $4N$-dimensional vector of ones, $\boldsymbol{\tau}$ is the total torque vector in the local body frame. The vector $\boldsymbol{\tau}$ is obtained by computing the thrust and torques developed by each quadcopter (from (4) and (5) in \cite{wang2016dynamics}) and translating them to the local body reference frame through the known relative position vectors of the attachment points.}

We linearize the rigid body dynamics model about hover configuration and decompose the inputs in $\mathbf{u} = \Bar{\mathbf{u}} + \mathbf{u'}$, in which $\Bar{\mathbf{u}}$ are the feedforward thrusts and $\mathbf{u}'$ are the first order components of the linearized model. We do the same for the state vector $\mathbf{x} = \Bar{\mathbf{x}} + \mathbf{x}'$. We model the disturbances applied to the system as random Gaussian noise in the form $\mathbf{d} = [\mathbf{f}_d^T, \mathbf{t}_d^T]^T$, in which $\mathbf{f}_d \in \mathbb{R}^3$ and $\mathbf{t}_d \in \mathbb{R}^3$ represent disturbance force and a torque applied to the origin of the local body reference frame (shown in Fig. \ref{fig:dronescheme}b). 
%

We can write the continuous-time linearized dynamics of the quadcopter-payload system (as a whole) as: 
\begin{equation}
    \Dot{\mathbf{x}}' = A \, \mathbf{x}' + B(\boldsymbol{\theta})\, \mathbf{u}' + B_d(\boldsymbol{\theta}) \, \mathbf{d}.
    \label{eq:dynamics}
\end{equation}
%
The dynamics matrix $A \in \mathbb{R}^{12\times12}$ is obtained \blue{by taking the Jacobian of \eqref{eq:nonlinear-dynamics} around hovering conditions with respect to the state vector $\mathbf{x}$}. $A$ does not depend on the placement variables $\boldsymbol{\theta}$. The matrices $B(\boldsymbol{\theta}) \in \mathbb{R}^{12\times4N}$ and $B_d(\boldsymbol{\theta})\in \mathbb{R}^{12\times6}$ \blue{are obtained by taking the Jacobian of \eqref{eq:nonlinear-dynamics} with respect to the input $\mathbf{u}$ and disturbance $\mathbf{d}$ vectors, respectively}. These matrices depend on $\boldsymbol{\theta}$ directly or through the inertia of the system (which itself is a function of $\boldsymbol{\theta}$). \blue{It is also important to note that the error due to the linearization of the dynamics equations does not depend on $\boldsymbol{\theta}$, and the linear approximation is reasonably accurate for the applications of interest in this paper.}
\vspace{-0.3cm}
\subsection{Control and layout co-design}
In this section, we present our co-design approach. First, we show how to derive the optimal $\mathcal{H}_2$ controller for the system of interest, and then how to use this approach to maximize robustness with respect to the layout variables. \blue{An alternative approach could be the $\mathcal{H}_\infty$ control \cite{dullerud2013course}, which by design is optimal for the worst-case disturbance. However, we find that $\mathcal{H}_2$ control is a better choice in our case. In fact, the worst-case disturbance for the systems of interest do not match the ones that are actually seen at deployment, which are better approximated as white noise.} 

With reference to the $\mathcal{H}_2$ optimal control literature \cite{dullerud2013course}, we define an additional auxiliary variable 
\begin{equation}
    \mathbf{z} = C \mathbf{x}' + D \mathbf{u}'.\label{eq:z}
\end{equation}
%
The goal is to find the feedback gain $K^*\in\mathbb{R}^{4N\times12}$ (dependent on $\boldsymbol{\theta}$) such that 
\begin{equation}
\mathbf{u}' = - K^*(\boldsymbol{\theta})\mathbf{x}', \label{eq:feedback}
\end{equation}
%
and minimize the integral $\mathcal{H}_2$ control cost $J = \int_0^\infty  \mathbf{z}^T\mathbf{z} \, dt$ when the system is subject to the Gaussian disturbance $\mathbf{d}$. \blue{In other words, the goal here is to capture the first-order effect of disturbances such as random forces and torques due to a wind gust. The linear approximation allows us to capture this first-order effect in a computationally tractable way, thanks to the following analytic derivation. This derivation would be less suitable for higher-order nonlinear metrics aiming at capturing, for example, the system maneuverability or maximum achievable speed.}

It is a known result that the optimal linear feedback controller minimizing the cost function $J$ is obtained as
\begin{equation}  K^*(\boldsymbol{\theta}) = (D^TD)^{-1} B(\boldsymbol{\theta})^T S_1(\boldsymbol{\theta}),
\end{equation}
where $S_1(\boldsymbol{\theta})$ is the solution of the algebraic Riccati equation
\begin{equation}
    A^T S_1 + S_1 A - S_1 B (D^T D)^{-1} B^T S_1 + C^T
C = 0 \label{eq:riccati1}.
\end{equation}
We omitted the dependency on $\boldsymbol{\theta}$ in \eqref{eq:riccati1} for clarity of writing. The feedback gain does not depend on the matrix $B_d$, which represents how the disturbance affects the dynamics. The optimal cost value is
\begin{equation}
    J^* = \text{trace}(B_d^T S_1 B_d),
\end{equation}
which instead does depend on the matrix $B_d$. This can be used to optimize the quadcopters' attachment locations. However, this formulation only considers feedback and ignores feedforward terms, and does not guarantee that the inputs generated by \eqref{eq:feedback} are feasible. Therefore, we do an additional step: we compute the closed-loop state covariance $S_2(\boldsymbol{\theta})$ of the system controlled with the optimal feedback defined in \eqref{eq:feedback}, subject to the same white noise disturbance $\mathbf{d}$. This can be computed using the optimal continuous-time state observer theory, which leads to the equation:
\begin{equation}
    A_f S_2 + S_2 A_f^T + B_d B_d^T = 0, \label{eq:riccati2}
\end{equation}
where we defined $A_f(\boldsymbol{\theta}) = A - B(\boldsymbol{\theta}) K^*(\boldsymbol{\theta})$ as the closed-loop dynamics matrix using the linear feedback of \eqref{eq:feedback}. Note that the Gaussian noise disturbance makes the process ergodic, and therefore the state sample mean is equivalent to its temporal mean. The matrix $S_2(\boldsymbol{\theta})$ is then an approximation of the covariance of the first order state linearization $\mathbf{x}'$ when subject to white noise disturbance. 

At this point, we have the feed-forward thrust values, which are computed from the hovering equilibrium condition, and the covariance matrix of the inputs $\Sigma_{\mathbf{u}} = K^*(\boldsymbol{\theta})^T S_2(\boldsymbol{\theta}) K^*(\boldsymbol{\theta})$, obtained through the feedback law. The goal is to maximize the probability of feasible inputs. We set the element-wise thrust lower saturation to be $\mathbf{u}_l = u_l \mathbf{1}_{4N}$ and the upper saturation to be $\mathbf{u}_h = u_h \mathbf{1}_{4N}$, where $\mathbf{1}_{4N}$ indicates the $4N$-dimensional vector of ones. We can then write the optimization problem we are trying to solve as:
\begin{equation}
    \max_{\boldsymbol{\theta}} \bigl\{ F(\mathbf{u}_h) - F(\mathbf{u}_l) \bigr\}
\end{equation}
\begin{figure}[!tb]
    \centering
    \includegraphics[width=0.4\textwidth]{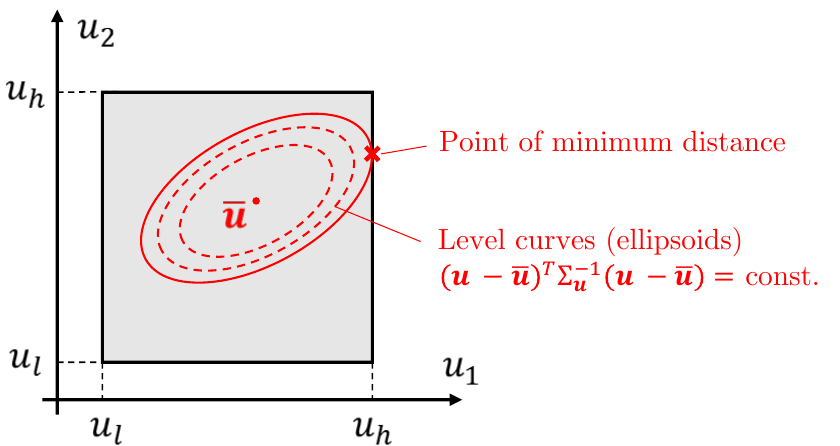}
    \caption{Two dimensional example of minimum Mahalanobis distance to input saturations. In this case, the random variables are $u_1, u_2 \in [u_l, u_h]$. The ellipses depicted, centered at the mean $\Bar{\mathbf{u}}$, represent the level curves of the Mahalanobis distance and are obtained as $(\mathbf{u} - \Bar{\mathbf{u}})^T\Sigma_\mathbf{u}^{-1}(\mathbf{u} - \Bar{\mathbf{u}}) = q$ ($q$ being a positive scalar parameter). The ``point of minimum distance'' sought is given by the tangency between the largest ellipsoid contained in the bounding box and the corresponding saturation hyperplane.}
\label{fig:mahalanobis}
    \vspace{-0.4cm}
\end{figure}
\hspace{-0.15cm}where $F(\cdot)$ is the cumulative distribution over thrust inputs, and its mean and covariance depend on $\boldsymbol{\theta}$. It is possible to approximate this cumulative distribution through sampling, but the sampling approach would lead to a noisy cost function for our optimization problem, whose variance decreases linearly with the number of samples used. 
In addition, sampling-based methods are computationally expensive and lead to long solution times. Therefore, we choose to leverage the concept of Mahalanobis distance \cite{mahalanobis2018generalized} to quantify the margin between mean (feedforward thrusts) and the thrust saturations. 
The Mahalanobis distance of a point $\mathbf{u}$ to a distribution with mean $\Bar{\mathbf{u}}$ and covariance $\Sigma_{\mathbf{u}}$ is defined as:
\begin{equation}
    d_M(\mathbf{u}; \Bar{\mathbf{u}}, \Sigma_{\mathbf{u}}) = \sqrt{(\mathbf{u} - \Bar{\mathbf{u}})^T \Sigma_{\mathbf{u}}^{-1} (\mathbf{u} - \Bar{\mathbf{u}})}. \label{eq:mahalanobis}
\end{equation}
We extend the concept of Mahalanobis distance of a hyperplane from a distribution as the minimum Mahalanobis distance of a point of the hyperplane from that distribution. A two dimensional example is shown in Fig. \ref{fig:mahalanobis}. Since in our case we have a lower and an upper saturation ($u_l$ and $u_h$ respectively) for each input component, and four thrust inputs per quadcopter, there are $8 N$ different hyperplanes to consider in the input space. Therefore, going back to the goal of maximizing the probability of feasible inputs, we aim to maximize the Mahalanobis distance of the thrust saturation hyperplanes from the input distribution. The optimization problem can be formulated as:
\begin{equation}
\begin{aligned}
    \max_{\boldsymbol{\theta}} \biggl\{\,\, \min_\mathbf{u}& \biggl[\,\, d_M\Bigl(\mathbf{u}; \Bar{\mathbf{u}}(\boldsymbol{\theta}), \Sigma_{\mathbf{u}}(\boldsymbol{\theta})\Bigl)\,\,\biggl] \,\,\biggl\} \\
    & \text{s.t.} \,\,\mathbf{u}_l \leq \mathbf{u} \leq \mathbf{u}_h \,\,\, \text{(element-wise)} \\
\label{eq:optimization}
\end{aligned}
\vspace{-0.5cm}
\end{equation}
%
We are looking for the value $\boldsymbol{\theta}^*$ which maximizes the minimum Mahalanobis distance of the thrust saturation hyperplanes from the feedforward thrust. This minimum distance is computed by evaluating it on each saturation hyperplane separately, hence this minimization step is component-wise and performed over both saturations $\{u_l, u_h\}$. The inner minimization over $\mathbf{u}$ is necessary because, on each hyperplane, we are interested in finding the point with minimum distance from $\Bar{\mathbf{u}}$. The optimization routine is also described in Algorithm \ref{alg:optim}. With the superscript expression $u^{(k)}$ we refer to the $k$-th component of the vector $\mathbf{u}$, while the expression $u_a \in \{u_l, u_h\}$ is used to denote which of the two thrust saturation constraints is \textit{active} at a given stage of the optimization. 

On the computational cost level, there are no particularly expensive operations. Solving Riccati equations is relatively fast, and minimizing the Mahalanobis distance from a hyperplane is a Quadratic Program (QP). For each update of $\boldsymbol{\theta}$ in the optimization routine, $8N$ QPs are solved. The solver used for the outer optimization loop is the Nelder-Mead simplex method, details of which can be found in \cite{nelder1965simplex} and \cite{lagarias1998convergence}.
In general, this optimization problem is non-convex, even if the payload shape is convex. A trivial example is the invariance to permutation of the $\boldsymbol{\theta}$ vector, which alters the optimizer but not the optimum. There are also cases in which local minima exist. To make sure the minimum is found in practice, we run the algorithm multiple times with different initial guesses.

An important observation to make is that our algorithm deals with feedback and feedforward in a joint fashion, maximizing the margin from actuator saturation when the system is in closed-loop. \blue{The optimal LQR controller is re-computed at each iteration and is thus a byproduct of our co-design tool.}
\begin{algorithm}[tb]
\caption{$\mathcal{H}_2$ based quadcopter placement optimization}
\begin{algorithmic}
    \Require $N, \text{quadcopters inertia}, \text{object inertia}, \text{object shape}$
    \Require $\text{weight matrices}\,\, C \,\,\text{and} \,\,D$
    \State Initial guess for $\boldsymbol{\theta}$
    \While{Optimization not completed}
        \State Compute overall system inertia
        \State Compute $A$, $B(\boldsymbol{\theta})$, $B_d(\boldsymbol{\theta})$ and feedforward $\Bar{\mathbf{u}}$
        \State Compute $S_1(\boldsymbol{\theta})$ and $K^*(\boldsymbol{\theta})$ from \eqref{eq:riccati1}
        \State Compute $S_2(\boldsymbol{\theta})$ from \eqref{eq:riccati2}
        \State Compute input covariance as $\Sigma_{\mathbf{u}} \gets K^{*T} S_2 K^*$
        \State Initialize $d_M^*$
        \For{each thrust input component $k$}
        \For{$u_a$ in $\{u_l, u_h\}$}
            \State Solve QP (as in \eqref{eq:optimization}):\\ \,\,\,\,\,\,\,\,\,\,\,\,\,\,\,\,\,\,\,\,\,\,\,\,\,\,\,\,\,\,\,\,\,\,\,$d' \gets \min_{\mathbf{u}} d_M(\mathbf{u};\Bar{\mathbf{u}}, \Sigma_{\mathbf{u}})$\\ \,\,\,\,\,\,\,\,\,\,\,\,\,\,\,\,\,\,\,\,\,\,\,\,\,\,\,\,\,\,\,\,\,\,\,\,\,\,\,\,\,\,\,\,\,\,\,\,\,\, s.t. $\mathbf{u}_l \leq \mathbf{u} \leq \mathbf{u}_h$, $u^{(k)} = u_a$
            \If{$d' < d_M^*$}
                \State $d_M^* \gets d'$
            \EndIf
        \EndFor
        \EndFor
        \State Update $\boldsymbol{\theta}$ (Nelder-Mead simplex algorithm)
    \EndWhile
\end{algorithmic}
\label{alg:optim}
\end{algorithm}
\vspace{-0.2cm}
\subsection{Control Implementation}
\begin{figure}[tb]
    \centering
    \includegraphics[width=0.49\textwidth]{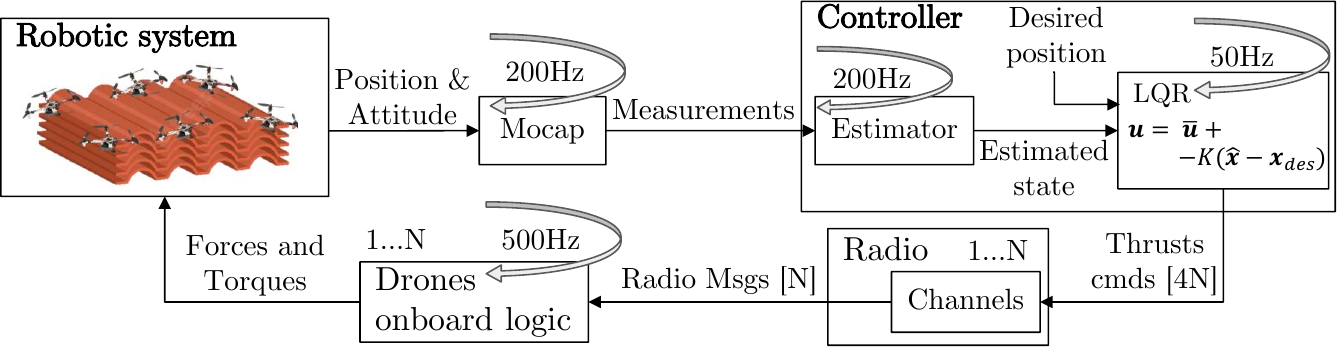}
    \caption{Diagram of the control infrastructure. The specific loop rates are relative to the available experimental testbed.}
    \label{fig:control}
\end{figure}

%
For simplicity, we treat the integrated quadcopter-payload system as a single rigid body, and use the LQR gains computed in the co-design optimization algorithm \ref{alg:optim}, \blue{which are readily available as a byproduct of our co-design tool.} 
This approach leads to a hierarchical control infrastructure, in which a single estimator is used for the overall motion and the quadcopters receive thrust commands individually through $N$ parallel communication channels. 
A block diagram of the control infrastructure is shown in Fig. \ref{fig:control}. 

\blue{An alternative control infrastructure, allowing compatibility with off-the-shelf autopilots, could be commanding a specific total thrust for each quadcopter. This approach would however be strictly suboptimal compared to independently commanding each motor.}
\blue{More advanced control strategies retaining the same first-order properties achieved by the $\mathcal{H}_2$ controller could also be used. One example could be a Model Predictive Control approach using the same optimal LQR weights as our co-design tool. Using alternative controllers is beyond the co-design focus of this work.}

The overall behavior of the controller is dominated by the choice of some hyperparameters which directly impact the computation of the $\mathcal{H}_2$ cost. First, the disturbance matrix $B_d$ from \eqref{eq:dynamics}, that describes how the random disturbance $\mathbf{d}$ affects the dynamics, and, in particular, quantifies its covariance along different directions in the wrench space. Second, the matrices $C$ and $D$ from \eqref{eq:z}, i.e.\ how the auxiliary variable $z$ is defined. For simplicity, the $B_d$ matrix was chosen to weigh the six disturbance components equally, i.e.\ $B_d(4,1) = B_d(5,2) = B_d(6,3) = B_d(10,4) = B_d(11,5) = B_d(12,6) = 1$. The $C$ matrix is chosen to just weigh the position of the payload centroid and the system's yaw angle (i.e.\ $C \in \mathbb{R}^{(4+4N)\times12}$). After evaluating the system's behavior in simulation, we set $C(1,1) = C(2,2) = 0.5$ (corresponding to the position on a horizontal plane $x-y$), $C(3,3) = 10$ (corresponding to vertical position $z$), $C(4,9) = 50$ (corresponding to yaw angle), and zero for all the other components. The input weight matrix $D \in \mathbb{R}^{(4+4N)\times4N}$ was set to the block matrix $D = [\mathbf{0}_{4N\times4}, I_{4N}]^T$, where $\mathbf{0}_{m\times n}$ denotes the $m\times n$ matrix of all zero elements and $I_{k}$ the $k$-dimensional identity.
\vspace{-0.3cm}
\subsection{Optimization Results \label{sec:results}}
\begin{figure}[tb]
    \vspace{-0.2cm}
    \centering
    \includegraphics[width=0.49\textwidth]{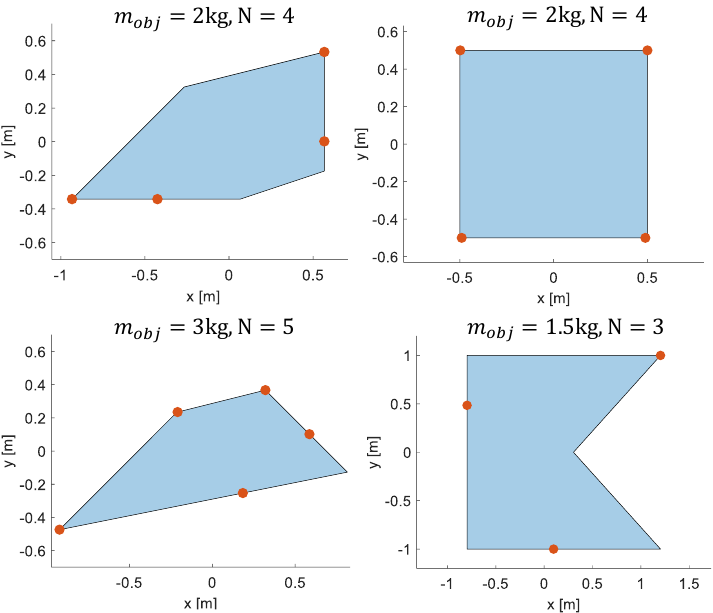}
    \caption{Optimal layouts for various object shapes and masses using different number of quadcopters. Our approach can handle both convex and non-convex polygons. It is interesting to note that, in contrast to common intuition, the optimal drone placement around a symmetric shape might not be symmetric.}
    \label{fig:results}
    \vspace{-0.4cm}
\end{figure}

The optimization framework we developed allows us to determine the optimal placement of quadcopters for transportation across different payload shapes, mass values and number of quadcopters used. In Fig. \ref{fig:results} we show some examples of optimal placements using the vehicles described in Tab. \ref{tab:params} as a reference. It is interesting to observe that the quadcopters are placed in a way which seems to trade off two different contributions. On the one hand, the area of the polygon defined by the quadcopters' attachment points tends to be maximized in order to maximize control authority. On the other hand the distances of the attachment points to the system's center of mass tend to be equalized. This is an effect of employing the Mahalanobis distance as a metric. In fact, even if the control authority is maximized (i.e.\ the input covariance $\Sigma_{\mathbf{u}}$ has smaller eigenvalues), we also need the quadcopters to have a comparable margin between feedforward thrusts $\Bar{\mathbf{u}}$ and saturation, otherwise the Mahalanobis distance objective is decreased.
Another interesting observation is that a symmetric object does not lead to a symmetric optimal configuration necessarily, and sometimes the optimum can be counterintuitive. This is shown by the concave shape in Fig. \ref{fig:results}.
\begin{table}[tb]
    \normalsize
    \caption{Table of quadcopter parameters}
    \resizebox{\linewidth}{!}{%
    \begin{tabular}{|c|c|c|c|}
    \hline
    quadcopter size (motor to motor) & $330\,\mathrm{mm}$ & propeller diameter & $203\,\mathrm{mm}$ \\
    \hline   
     quadcopter frame mass & $525\,\mathrm{g}$ & battery mass & $437\,\mathrm{g}$\\
    \hline
    \end{tabular}%
    }
    \label{tab:params}
    \vspace{-0.4cm}
\end{table}
\vspace{-0.1cm}
\section{Experiments\label{sec:experimental}}
The relevant parameters of the vehicles used are shown in Tab. \ref{tab:params}.
The battery capacity ($5300\,\mathrm{mAh}$) has been chosen to guarantee $10$ minutes of flight time at maximum payload. The system theoretically is able to lift an external payload of up to $\sim 0.8\,\mathrm{kg}$ per quadcopter. It is interesting to note that the main contribution to the overall inertia of the system is due to the quadcopters, since they are attached to the payload perimeter, and thus at a larger distance from the overall center of mass. It is also important to highlight that contributing a greater inertia than the actual payload is in line with all the ground and aerial transportation systems in use nowadays. In fact, most of the transportation systems that we use daily (e.g.\ trains, planes, cars) have significantly greater mass and inertia compared to the payload they transport.

\begin{figure}[tb]
    \centering
    \includegraphics[width=0.49\textwidth]{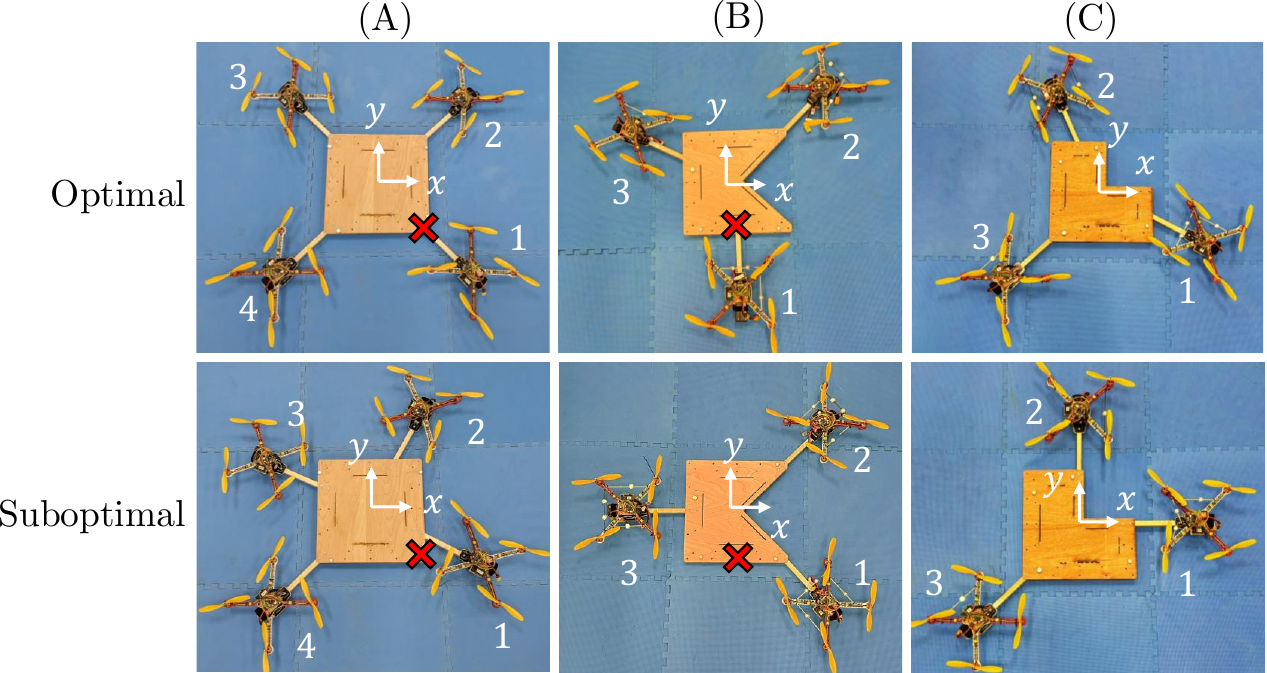}
    \caption{\blue{Attachment layouts tested on panel A, B, and C. Top row: optimal configurations, bottom row: suboptimal configuration. The red cross indicates where the disturbance mass is attached in the disturbance rejection experiments on A and B. Panel C is used for trajectory tracking experiments.}}
    \label{fig:configs}
    \vspace{-0.4cm}
\end{figure}

For the flight experiments \blue{three different payload shapes were used : a square panel (A) of mass $1.02\,\mathrm{kg}$, a concave square (B) of mass $0.71\,\mathrm{kg}$, and an L-shaped panel of mass $0.76\,\mathrm{kg}$, all of side length $0.45\,\mathrm{m}$. For these three options, the performance of optimal quadcopter configurations were tested against some suboptimal case (see Fig. \ref{fig:configs}). In the case of Panel B, a symmetric layout was chosen as the suboptimal configuration. Panel A was lifted by four quadcopters, while for panel B and C only three were used. On panels A and B, we performed hovering tests (with and without wind), response to position reference step tests, and disturbance rejection tests when suddenly attaching a disturbance mass.
Panel C was used for a trajectory tracking experiment.}
\blue{Our experiments aim at showing the \textit{comparative} advantages achieved through the proposed method, rather than focusing on the absolute performance metrics of the controller, which can be optimized through various other state-of-the-art methods.}

The rigid attachments were obtained by directly bolting the quadcopters to the panels through wood dowels. The experiments are conducted in an indoor flight space equipped with a Motion Capture system, and markers for motion tracking are placed directly on the panels. A picture of the system flying in the flight space is shown in Fig. \ref{fig:system_photo}.

\vspace{-0.005cm}
The square panel (A) was also used to test the payload capacity of the system. Four quadcopters attached at the corners were able to lift, hover and land with up to $\sim2.5\,\mathrm{kg}$. Large payload masses, however, are not ideal due to the narrow margin from actuator saturations. The reduced control authority causes the system to have limited disturbance response capabilities, especially within the spatial limits of the flight space available for experiments.
Supplementary videos showing relevant experiments can be found at \texttt{\url{https://hiperlab.berkeley.edu/members/carlo-bosio/automatedlayout/}}.
%
\vspace{-0.3cm}
\subsection{Hovering}
As a first comparison, we report the hovering performances of the different systems in the various configurations we tested. The hovering was performed in still air and with a $1\,\mathrm{m \cdot s^{-1}}$ wind disturbance. The RMSE values, averaged across five runs roughly $15$ seconds long, of position deviation from reference and attitude angles can be found in Tabs. \ref{tab:hover} and \ref{tab:wind}. The optimal configurations for both panels are more stable and less subject to disturbance. The difference is particularly significant for Panel (B), which could not withstand the wind disturbance at all, and showed an unstable behavior (due to actuator saturation) in all the five trials.

\blue{Real world systems are inevitably affected by noise (e.g., imperfect sensors, communication delays, small modeling inaccuracies). Additionally, due to the sizes of the system ($\sim 1.5\,\mathrm{m}$) compared to the size of the flight space ($\sim 5\,\mathrm{m}$), significant undesired aerodynamic perturbations are present. These disturbances are unavoidable and partially explain the imperfect control absolute performances. However, they are present in the same amount regardless of the specific drone configuration. The comparative analysis suggests that the configuration plays an important role in improving the disturbance robustness.}

\blue{We also compare the hovering performances of an $\mathcal{H}_\infty$ controller on the optimal configuration for Panel (B) (Fig. \ref{fig:configs}). We simulate the system hovering subject to the nominal Gaussian noise for 10 minutes (for convergence of the RMSE statistics). The RMSE values are at least $50\%$ lower for the nominal $\mathcal{H}_2$ controller. We experimentally verify this by testing the system, controlled through $\mathcal{H}_\infty$, hovering in presence of $1 \mathrm{m\cdot s^{-1}}$ wind. The experiments confirm the higher performances of the $\mathcal{H}_2$ control with similar differences in the RMSE values. This is expected as the worst-case disturbances considered by $\mathcal{H}_\infty$ synthesis are likely not observed in this setting. In a scenario like ours, where the disturbance is reasonably modeled as white noise, the $\mathcal{H}_2$ approach leads to higher performance. Arguments and results of similar nature can be found in the literature \cite{siang2012comparison, shukla2016study}.}
%
\begin{table}[tb]
    \normalsize
    \begin{tabular}{|c|c|c|c|c|c|c|}
    \hline
    \multicolumn{7}{|c|}{No wind}\\
    \hline   
     Panel & \multicolumn{3}{|c|}{(A)} & \multicolumn{3}{|c|}{(B)}\\
     \hline
      Config. & S-opt. & Opt. & $\%$ & S-opt. & Opt. & $\%$ \\
      \hline
     $x\,[\mathrm{m}]$ & 0.284 & 0.1390 & \textbf{51} & 0.320 & 0.0547 & \textbf{83}\\
     \hline
     $y\,[\mathrm{m}]$ & 0.1440 & 0.0775 & \textbf{46} & 0.1240 & 0.0909 & \textbf{27}\\
     \hline
     $z\,[\mathrm{m}]$ & 0.0713 & 0.0674 & \textbf{5.5} & 0.0690 & 0.0619 & \textbf{10}\\
     \hline
     $Y\,[\mathrm{rad}]$ & 0.0547 & 0.0537 & \textbf{1.8} & 0.0687 & 0.0608 & \textbf{12}\\
     \hline
     $P\,[\mathrm{rad}]$ & 0.0235 & 0.0163 & \textbf{31} & 0.0231 & 0.0146 & \textbf{37}\\
     \hline
     $R\,[\mathrm{rad}]$ & 0.0213 & 0.0149 & \textbf{30} & 0.0206 & 0.0161 & \textbf{21}\\
    \hline
    \end{tabular}%
    \caption{Table of RMSE values for position and attitude averaged across five different undisturbed hovering trials for panels (A) and (B), with the respective suboptimal (S-opt. column) and optimal (Opt. column) configurations. The relative improvement of performances can be seen in the bold columns, in percentage.}
    \label{tab:hover}
    \vspace{-0.3cm}
\end{table}
\begin{table}[ht]
    \normalsize
    \begin{tabular}{|c|c|c|c|c|c|c|}
    \hline
    \multicolumn{7}{|c|}{$1\,\mathrm{m\cdot s^{-1}}$ wind}\\
    \hline   
     Panel & \multicolumn{3}{|c|}{(A)} & \multicolumn{3}{|c|}{(B)}\\
     \hline
      Config. & S-opt. & Opt. & $\%$ & S-opt. & Opt. & $\%$ \\
      \hline
     $x\,[\mathrm{m}]$ & 0.6374 & 0.216 & \textbf{66} & N/A & 0.1285 & -\\
     \hline
     $y\,[\mathrm{m}]$ & 0.1520 & 0.1373 & \textbf{9.6} & N/A & 0.1144 & -\\
     \hline
     $z\,[\mathrm{m}]$ & 0.5203 & 0.465 & \textbf{10.5} & N/A & 0.0773 & -\\
     \hline
     $Y\,[\mathrm{rad}]$ & 0.1168 & 0.0890 & \textbf{24} & N/A & 0.1031 & -\\
     \hline
     $P\,[\mathrm{rad}]$ & 0.0257 & 0.0182 & \textbf{29} & N/A & 0.0282 & -\\
     \hline
     $R\,[\mathrm{rad}]$ & 0.0232 & 0.0156 & \textbf{33} & N/A & 0.0407 & -\\
    \hline
    \end{tabular}%
    \caption{Table of RMSE values for position and attitude averaged across five different hovering trials when a $1\,\mathrm{m\cdot s^{-1}}$ wind is acting on the system. Data are reported for panels (A) and (B), with the respective suboptimal (S-opt. column) and optimal (Opt. column) configurations. The relative improvement of performances can be seen in the bold columns, in percentage. \blue{The suboptimal configuration for panel (B) was not able to withstand the wind disturbance, due to thrust saturations.}}
    \label{tab:wind}
    \vspace{-0.2cm}
\end{table}
%
\vspace{-0.2cm}
\subsection{Reference Step}
\begin{figure}[tb]
    \centering
    \includegraphics[width=0.4\textwidth]{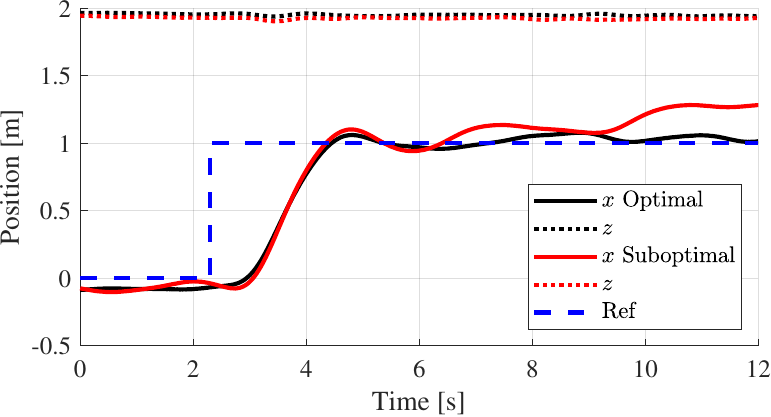}
    \caption{$x-z$ positions of the optimal and suboptimal configurations on Panel (A). The position reference step consists of an offset of $1\,\mathrm{m}$ along $x$. The optimal configuration has lower overshoot and faster recovery.}
    \label{fig:refstep}
    \vspace{-0.4cm}
\end{figure}

As a second performance test, with the system hovering, a reference step change of $1\,\mathrm{m}$ is applied, and the transient response is recorded. In Fig. \ref{fig:refstep} the response of the square Panel (A) system to an applied position reference step (in particular a $1\,\mathrm{m}$ offset along the $x$ direction) is shown. It can be observed that the optimal configuration has a lower overshoot, but also a faster transient. After six seconds, the suboptimal configuration has not yet fully recovered from the event.
\vspace{-0.2cm}
\subsection{Trajectory tracking}
\begin{figure}[tb]
    \centering
    \includegraphics[width=0.99\linewidth]{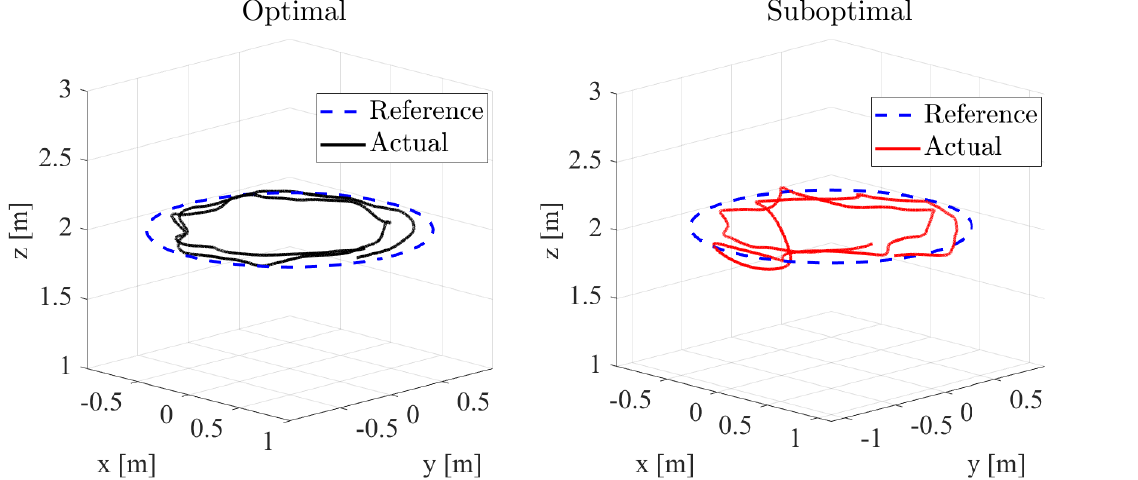}
    \caption{\blue{Example trajectory tracking experiments using the L-shaped panel (C). The commanded trajectory is a circle of diameter $2\,\mathrm{m}$ within the horizontal plane of height $z=2\,\mathrm{m}$.}}
    \label{fig:L-circ-traj}
    \vspace{-0.2cm}
\end{figure}

\blue{We also conducted trajectory tracking experiments on the L-shaped panel (C in Fig. \ref{fig:configs}). The commanded trajectory was a circle of diameter $2\,\mathrm{m}$ within the horizontal plane of height $z=2\,\mathrm{m}$. The reference velocity is $0.5\,\mathrm{m\cdot s^{-1}}$. Some example trajectories obtained are shown in Fig. \ref{fig:L-circ-traj}, which show substantial qualitative differences in the tracking performances.}
\vspace{-0.2cm}
\subsection{Disturbance Rejection}
As a third performance test, a disturbance mass was attached underneath the panels during flight, specifically at the attachment points shown by the red cross signs in Fig. \ref{fig:configs}. The mass attachment is done through a magnet to guarantee the repeatability of the experiment across different runs. Both the state and input data were collected and compared to validate their consistency with the layout optimization framework. For Panel (A), the mass was of about $0.5\,\mathrm{kg}$ (approximately $10\%$ of the total mass), whilst for Panel (B), it was of about $0.3\,\mathrm{kg}$ ($7\%$ of the total mass).
Once again, in the case of Panel (B), after a significant transient, most of the time the suboptimal configuration was not able to re-stabilize after the application of the disturbance. 

\begin{figure*}[tb]
    \centering
    \includegraphics[width=0.99\textwidth]{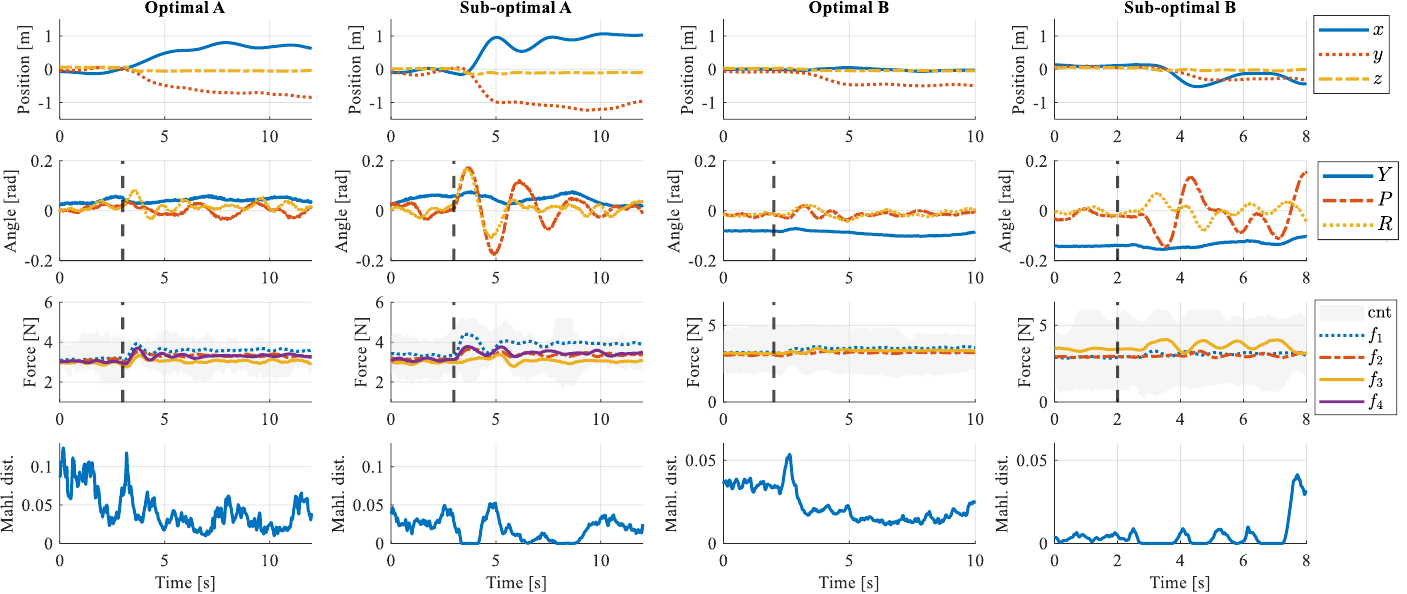}
    \caption{Response to step disturbance (added mass) over time. The mass is attached through a magnet, respectively where indicated by the red crosses in Fig. \ref{fig:configs}. The attachment instant is indicated with a vertical dashed line in the plots. On the top row: position deviation from set point. On the mid row: attitude response. On the bottom row: mean thrusts across each quadcopter (with dispersion bands in grey). Left: optimal and suboptimal configurations for panel (A), right: optimal and suboptimal configurations for panel (B). The grey area labeled as `cnt' in the bottom row represents the contours of the individual propeller thrusts (not averaged across each quadcopter). As it is possible to observe, the suboptimal configurations have wider contours. \blue{The bottom row shows the minimum Mahalanobis distance (defined in \eqref{eq:mahalanobis} and depicted in Fig. \ref{fig:mahalanobis}) from the saturations throughout the experiments (higher is better). The optimal placement of the quadcopters allows the system to overall maintain a greater normalized distance from the actuator saturations.}}
    \label{fig:plots}
    \vspace{-0.4cm}
\end{figure*}

Representative response plots over time are presented in Fig. \ref{fig:plots}.
From the data, it is clear that the choice of quadcopter layout has a significant impact on disturbance rejection performance. The optimal layout leads to lower oscillations in the attitude of the vehicle, specifically in the pitch and roll angles, and also recovers more quickly from disturbances. In comparison, the suboptimal layout sees its peak deviation from hovering conditions nearly double the one of the optimal setup. Another observation is related to thrust behavior after adding the payload. In the suboptimal configuration, the increase in thrust following the addition of the payload is more noticeable. Additionally, the steady-state thrust from the quadcopter positioned closest to the added payload is also higher in the suboptimal setup. This suggests that, despite facing disturbances of the same magnitude, the thrust distribution becomes more uneven in the suboptimal case, emphasizing the benefits of an optimal quadcopter layout. \blue{Finally, the bottom row of Fig. \ref{fig:plots} shows the minimum Mahalanobis distance (defined in \eqref{eq:mahalanobis} and depicted in Fig. \ref{fig:mahalanobis}) from actuator saturations. The optimal layouts, in general, maintain a greater distance from this unsafe condition.}
%
\vspace{-0.6cm}
\section{Conclusion\label{sec:conclusions}}
In this study, we introduced a novel approach to the layout and control co-design of cooperative aerial transportation systems. This methodology jointly solves the layout design and control synthesis problems through an optimization routine inspired from $\mathcal{H}_2$ control theory.
Our optimization framework allows to cater our system to different payload shapes, inertias, and quadcopter counts. 
The choice of the weight matrices $C$ and $D$ represents the freedom given to the control designer to obtain the desired performances.
\blue{Overall, the main contribution of this paper is a co-design tool using ideas from robust control to compute the optimal layout of thrust modules around a payload. The two main components of our approach are:
\begin{itemize}[leftmargin=*, noitemsep]
    \item A novel cost function able to capture the robustness to disturbances of the multi-UAV system;
    \item An efficient procedure to optimize this cost function with respect to the layout variables.
\end{itemize}}
We also emphasize the idea that dealing with physical parameters and control jointly increases the overall system performances compared to a sequential, siloed approach (control design following layout design), sometimes even making the difference between failure and success. \blue{Moreover, extending this to nonlinear controllers could be done as follows: a desired nonlinear control strategy is implemented, and the control gains are selected so that the linearized feedback matches the computed gains. Of course, arguments about saturation become weaker, as saturation is likely to occur in areas where linearization is no longer accurate.}

The experimental results validate the effectiveness of our approach\blue{, even when fine engineering effort are not made to estimate system parameters and develop sophisticated controllers}. Flight tests under disturbance highlighted the robustness of the system, and the agreement between the performance predicted by our optimization tool and experimental evidence.

Possible extensions for this work include exploring more advanced control strategies for the deployment stage
and doing online payload parameter estimation in cases of uncertainty.
%
To conclude, our study showed that the co-design of control and physical parameters in robotics is a promising path to achieve superior performance and reliability.
%
\vspace{-0.2cm}
\section*{Acknowledgments}
This work was supported by the Tsinghua-Berkeley-Shenzhen Institute (TBSI), and the Powley fund of the Mechanical Engineering department of UC Berkeley. The experimental testbed at the HiPeRLab is the result of contributions of many people, a full list of which can be found at \href{https://hiperlab.berkeley.edu/members/}{\texttt{hiperlab.berkeley.edu/members/}}.
\bibliography{IEEEabrv, refs}
\vfill
\end{document}